\documentclass{article}
\usepackage{graphicx} 

\usepackage[square,numbers]{natbib}
\usepackage{bbm}
\usepackage{amssymb}
\usepackage{amsmath}

\usepackage{hyperref}

\usepackage{appendix}

\newcommand{\keywords}[1]{\textbf{Keywords:} #1}

\title{Semantic Segmentation from Image Labels by Reconstruction from Structured Decomposition}
\author{Xuanrui Zeng}
\date{July 2024}

\begin{document}

\maketitle

\begin{abstract}
Weakly supervised image segmentation (WSSS) from image tags remains challenging due to its under-constraint nature. Most mainstream work focus on the extraction of class activation map (CAM) and imposing various additional regularization. Contrary to the mainstream, we propose to frame WSSS as a problem of reconstruction from decomposition of the image using its mask, under which most regularization are embedded implicitly within the framework of the new problem. Our approach has demonstrated promising results on initial experiments, and shown robustness against the problem of background ambiguity. Our code is available at \url{https://github.com/xuanrui-work/WSSSByRec}.
\end{abstract}


\section{Introduction}

In this work, we present a novel generative view on weak(ly supervised) segmentation from image tags along with its related experiments. Weak segmentation is an appealing direction moving forward of full supervision due to its significantly lower cost of labelling. However, it also inherently poses greater challenges than fully-supervised segmentation. For one, the problem is significantly under-constraint, requiring additional regularization to achieve good local optima. For another, the notion of "background" and "foreground" is inherently inconsistent and ambiguous across images, thus without full mask supervision, it is difficult for a neural network to obtain meaningful correlations between the two for good segmentation. We hereby present an attempt to tackle these two problems by framing the weak segmentation problem as a problem of reconstruction under constraints.

\section{Related works}

Our work is a continuum of previous works and inspirations are drawn from the works \cite{pathak2015constrained}, \cite{kolesnikov2016seed}, and \cite{araslanov2020single}. \cite{pathak2015constrained} proposes to solve weak segmentation via constrained optimization: instead of directly regularizing the output mask (distribution) of the network, they regularize it through the introduction of an intermediate latent distribution, of which various linear constraints on mask sizes can be directly incorporated. The output mask is then regularized to be close to the latent distribution. \cite{kolesnikov2016seed} divides weak segmentation into the joint optimization of three objectives: seeding, expansion, and contraction. In the seeding objective, the output mask is regularized to be consistent with the localization seed provided by some weak localization procedure such as class activation map (CAM) \cite{zhou2016learning}; the expansion objective places regularization on the mask sizes; and the contraction objective regularizes the output mask to be consistent with low-level image appearances by penalizing its difference to the mask outputted by an existing low-level segmentation algorithm, such as the CRFs \cite{krahenbuhl2011efficient}. \cite{araslanov2020single} proposes a different pipeline where the output mask is treated as the CAM with its global average pooled vector forming a classification loss, similar to the original CAM in \cite{zhou2016learning}. Aside from this, mask output from the network is refined using a separate iterative procedure and fed back into the network as pseudo-label for training as the training progresses.

Different from the above works, which focus primarily on various regularization on the output mask, we propose to treat weak segmentation as a problem of decompose-recompose the input image with structured latent variables, of which additional regularization tailored to each individual latent variable are added as needed.

\section{Method}

\begin{figure}
    \centering
    \includegraphics[width=1\linewidth]{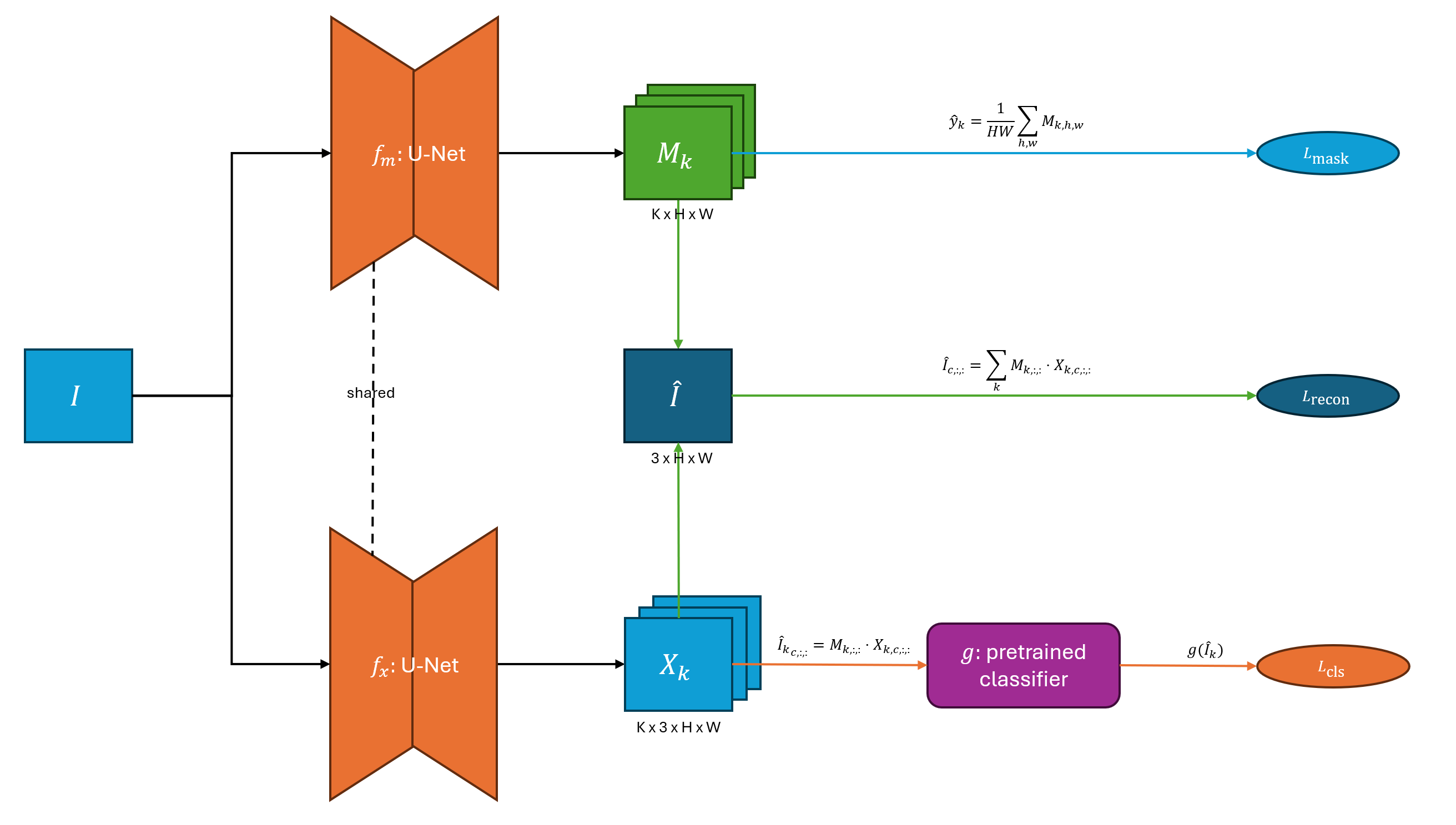}
    \caption{Overview of our pipeline for weak segmentation.}
    \label{fig:overview}
\end{figure}

In weak segmentation from image tags, we are given a dataset comprised of images and their respective tags indicating the presenting classes in each image, denoted by $D = \{ (I_i, y_i) \}_{i=1}^N$. $I_i \in \mathbb{R}^{3 \times H \times W}$ is the $i$-th RGB image in the dataset. $y_i \in \{0, 1\}^K$ is the label vector for $I_i$ with dimension $K$, the number of classes including the "background" class, and with each dimension $j$ of $y_i$, denoted by $y_{ij}$, containing the indicator variable of the presence of class $j$ in $I_i$. The goal of weak segmentation is to obtain a model $h: \mathbb{R}^{3 \times H \times W} \rightarrow [0, 1]^{K \times H \times W}$ that given an image $I$ outputs its corresponding segmentation mask $M = h(I)$, where $M_{:,h,w}$, represents the categorical distribution over classes at pixel location $(h,w)$ in image $I$.

Figure \ref{fig:overview} shows an overview of our approach. We frame the weak segmentation problem into an equivalent problem of learning a decomposition of the input image. More specifically, we aim to learn two neural networks: the mask network $f_m: \mathbb{R}^{3 \times H \times W} \rightarrow [0, 1]^{K \times H \times W}$ and the decomposition network $f_x: \mathbb{R}^{3 \times H \times W} \rightarrow \mathbb{R}^{K \times 3 \times H \times W}$. The mask network follows the same schema as $h$ mentioned above and outputs the mask $M = f_m(I)$. We denote the probability map for the $k$-th class in the mask $M_k$ and refer to it as "mask-let", which contains the probability of each of pixels in $I$ belonging to the $k$-th class. The decomposition network is novel and outputs a decomposition of the input image $I$ into a set of $K$ images $X = f_x(I) = \{X_k\}_{k=1}^K$, with each $X_k$ we refer to "image-let". Intuitively, we desire each $X_k$ to \textit{contain} the portions of $I$ that should be classified as class $k$, with the rest of the portions we ignore.

Under this setup, a reconstruction of $I$ is formed by re-composing image-lets $\{X_k\}$ using mask-lets $\{M_k\}$ as the weights, for which we denote $\hat{I}$:
\begin{align} \label{eq:recon}
    \hat{I}_{c,h,w} = \sum_k M_{k,h,w} \cdot X_{k,c,w,h} .
\end{align}

\subsection{Loss Functions}

Our revised objective for weak segmentation focuses on reconstructing the input image $I$ subject to additional soft-constraints placed on each $X_k$ and $M_k$ and is composed of three parts: the reconstruction loss $L_\text{recon}$, the mask constraint loss $L_\text{mask}$, and the image-let constraint loss $L_\text{cls}$.

\subsubsection{$L_\text{recon}$}

The reconstructed image $\hat{I}$ should be similar to $I$. Intuitively, this implies a sensible decomposition of the image into the respective mask-lets and image-lets. It can also be seen as a soft-constraint on $\{M_k\}$ and $\{X_k\}$ which specifies that their weighted-sum should be identical to $I$. For simplicity, we choose to use the squared-error to penalize dissimilarity between $\hat{I}$ and $I$, which give rises to the following form for $L_{recon}$:
\begin{align}
    L_\text{recon}(\hat{I}, I) = \frac{1}{CHW} \sum_{c,h,w} \Vert \hat{I}_{c,h,w} - I_{c,h,w} \Vert ^2 .
\end{align}

\subsubsection{$L_\text{mask}$}

Each mask-let $M_k$ should be reflective of the presence/absence of class $k$. For class $k$ present ($y_k = 1$), the average score over all pixels $\hat{y}_k$:
\begin{align}
    \hat{y}_k = \frac{1}{HW} \sum_{h,w} M_{k,h,w} ,
\end{align}
should be large, where as for class $k$ absent ($y_k = 0$), $\hat{y}_k$ should be small. To accomplish this, we use the following cross-entropy-like loss for $L_\text{mask}$:
\begin{align}
    L_\text{mask}(\hat{y}, y) = -\frac{1}{K} \sum_{k} y_k \log(\hat{y}_k) + (1 - y_k) \log(1 - \hat{y}_k) .
\end{align}
Note that our choice of $L_\text{mask}$ and $y$ implicitly places a prior encouraging equal mask size for the presenting object of each class, since $\hat{y}_k$ can be viewed as the normalized expected area for class $k$, and $L_\text{mask}$ weights the presenting classes equally by $y_k = 1$ in the sum. Additional beliefs on mask size can be incorporated straight-forwardly by changing the label vector $y$ into a soft-onehot vector, with each component $y_k$ denoting the groundtruth normalized area for class $k$ given the input image $I$ instead of an indicator variable.

\subsubsection{$L_\text{cls}$}

It is desirable for each image-let $X_k$ of the presenting class to include the portion relevant to class $k$ while excluding portions for other classes, which effectively translates into separation of image regions for different classes into different respective image-lets. We achieve this via gradient guidance from a pretrained multi-class classifier denoted by $g: \mathbb{R}^{C \times H \times W} \rightarrow [0,1]^{K-1}$ via the loss $L_\text{cls}$.

Using the same dataset $D$, $g$ is trained a priori to output a probability vector $\hat{z} = g(I)$ with each component $\hat{z}_k$ modeling the probability of class $k$ being present in $I$, except for the "background" class, hence $\hat{z} \in [0,1]^{K-1}$. Without loss of generality, we assume that the background class is indicated by the last component of the label vector, $y_K$.

We define $L_\text{cls}$ as follows:
\begin{align}
    \hat{I}_{k_{c,h,w}} &= M_{k,h,w} \cdot X_{k,c,h,w} \\
    L_\text{cls}(M, X) &= \frac{1}{K} \sum_{k} \left ( - \mathbbm{1} [y_k = 1 \land k \neq K]\log (g_k(\hat{I}_k)) - \sum_{j \in \{1,...K\} \setminus k} \log (1 - g_j(\hat{I}_k))  \right ) .
\end{align}
where $g_k(\cdot)$ denotes the score of class $k$ outputted by $g$, with the parameter of $g$ fixed throughout training, and $\hat{I}_k$ here is the component from class $k$ that contributed to the reconstruction $\hat{I}$ in Eq. \eqref{eq:recon}.

Crucially the indicator function above $\mathbbm{1} [k \neq K]$ causes the first term in the summation to be zero for the background, such that $L_\text{cls}$ penalizes the image-let of the background class $X_K$ to exclude any potential foreground objects identifiable by $g$. This in return solves for the ambiguity of the concept of background across images. Furthermore, we suggest that this also help with class imbalance related to the background class, as it is no longer sufficient for the network to blindly output a large mask for the background in return for lower loss, since this will likely cause lots of foreground objects to appear in $\hat{I}_k$ and causes activations on $g$.

\subsubsection{Overall Loss}

The overall loss $L$ for a single sample $(I, y) \in D$ is the weighted combination of the above losses, given by:
\begin{align}
    L(M, X, I, y) = L_\text{recon}(\hat{I}, I) + \lambda_m \cdot L_\text{mask}(\hat{y}, y) + \lambda_c \cdot L_\text{cls}(M, X) ,
\end{align}
where $\lambda_m$ and $\lambda_c$ are two respective hyperparameters controlling the weights of the respective loss. Our optimization objective is thus:
\begin{align}
    {\arg \min}_{f_m, f_x} \frac{1}{|D|} \sum_{(I_i, y_i) \in D} L(f_m(I_i), f_x(I_i), I_i, y_i) .
\end{align}

\section{Experiments}

\subsection{Dataset}

As a proof of concept, we derive a custom dataset for binary segmentation of dogs in the image from the ImageNet-1k dataset \cite{imagenet15russakovsky}. The dataset consists of 20,000 RGB images of size $224 \times 224$ in total, out of which 10,000 are randomly selected images labelled as any category of dogs in the ImageNet-1k, and the other 10,000 are randomly selected images that aren't labelled as dogs. The former are annotated with a label vector of $y_i = [1,1]$ indicating the presence of both dog ($y_1 = 1$) and background ($y_2 = 1$), whereas the latter are annotated with a label vector of $y_i = [0, 1]$ indicating the absence of dog. We use 16,000 of the images for training, and leave the remaining 4000 for validation.

\subsection{Architecture}

With regards to the neural network architecture, we choose to use the U-Net \cite{ronneberger2015u} for both $f_m$ and $f_x$, and used weight-sharing for the encoder portion of both U-Net. Specifics of the network architecture are available in Appendix \ref{app:network_arch}. The pretrained multi-class classifier $g$ is derived from a ResNet-18 \cite{he2016deep} pretrained on ImageNet-1k with its last linear layer replaced to a single neuron outputting the score of dog presenting in image.

\subsection{Training}

$g$ is trained first on the 16,000 images for 10 epochs using the Adam optimizer \cite{kingma2014adam} with learning rate of 0.0001, betas (0.9, 0.999), and batch size of 32, with the following multi-class classification loss on each sample:
\begin{align}
    L_g(g(I), y) = \sum_{k=1}^{K-1} -y_i \log(g_i(I)) - (1-y_i) \log(1 - g_i(I)) .
\end{align}

Next, $g$ is fixed, and $f_m$ and $f_x$ are trained jointly on the same 16,000 images for 10 epochs using the previously prescribed losses with hyperparameters $\lambda_m = 1.0 \times 10^{-3}, \lambda_c = 1.0 \times 10^{-3}$, and optimized using Adam with learning rate of 0.0001, betas (0.9, 0.999), and batch size of 4.

\subsection{Qualitative Results}

\begin{figure}
    \centering
    \begin{minipage}[t]{0.45\linewidth}
        \includegraphics[width=\linewidth]{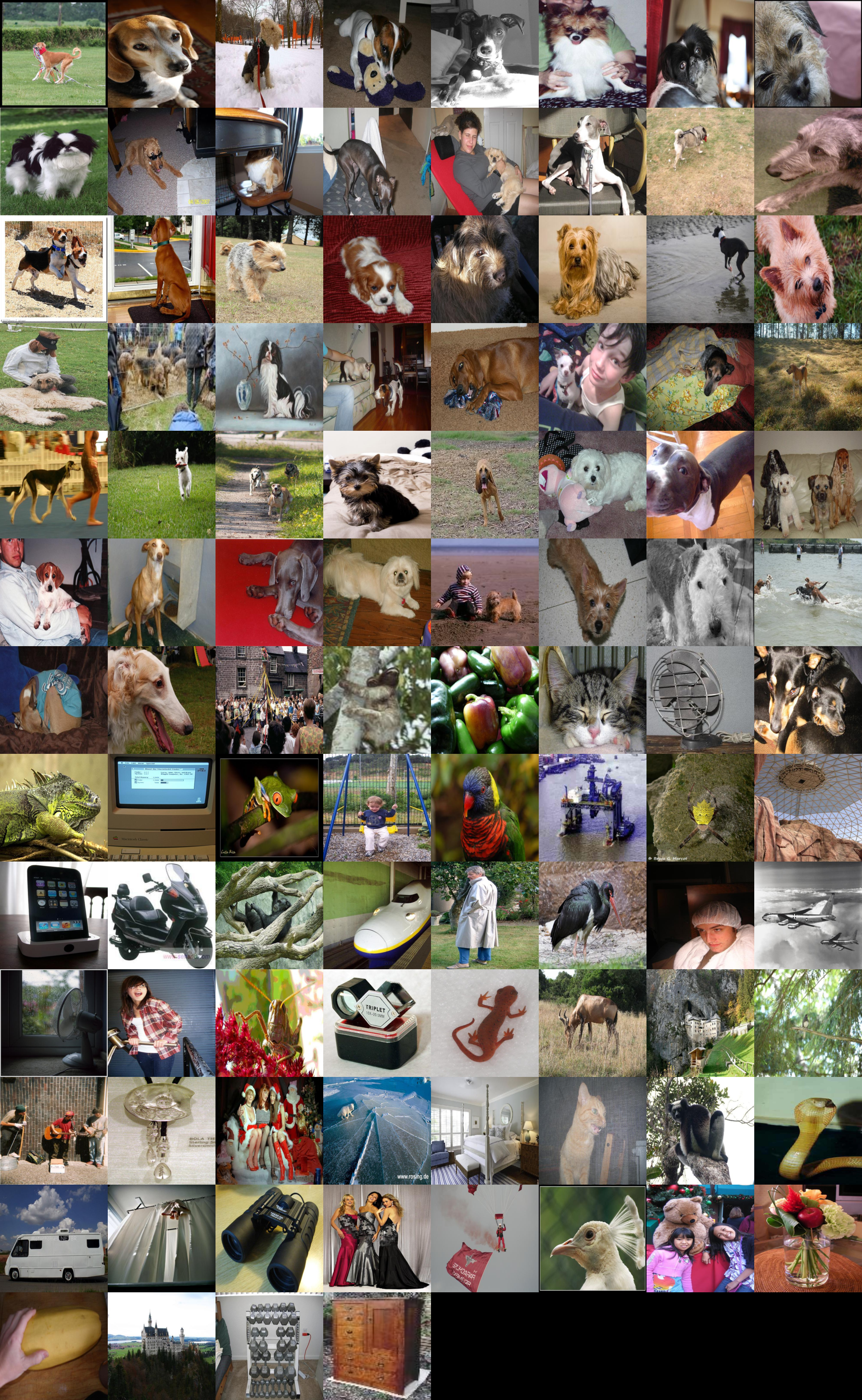}
    \end{minipage}
    \hspace{0.01\linewidth}
    \begin{minipage}[t]{0.45\linewidth}
        \includegraphics[width=\linewidth]{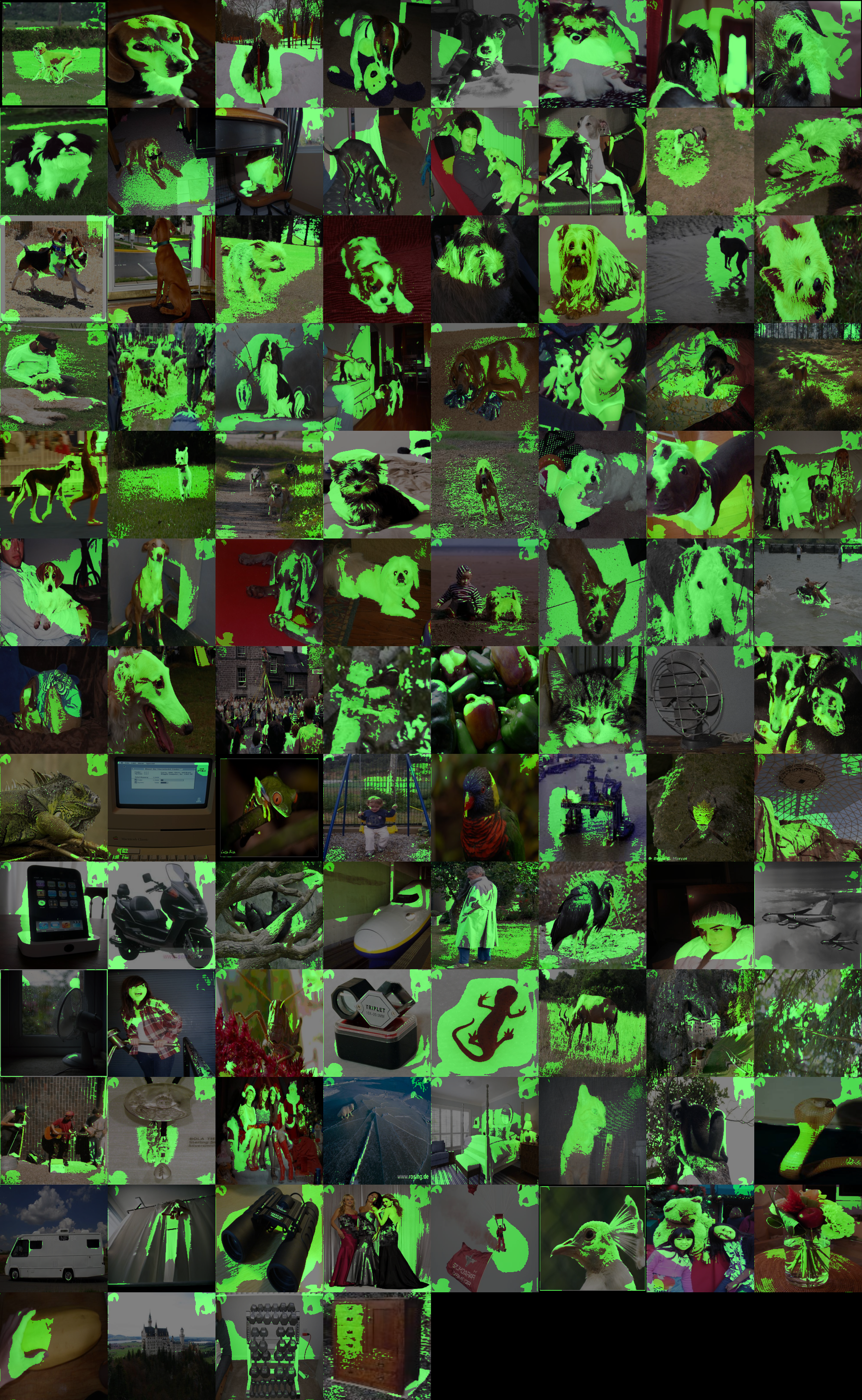}
    \end{minipage}
    \caption{Qualitative results on 50 randomly selected \textbf{training samples} for "dog-present" images (first 50) and "dog-absent" (last 50) images respectively. Left, each square: input image. Right, each square: output mask overlayed on the input image.}
    \label{fig:train_qua_results}
\end{figure}

\begin{figure}
    \centering
    \begin{minipage}[t]{0.45\linewidth}
        \includegraphics[width=\linewidth]{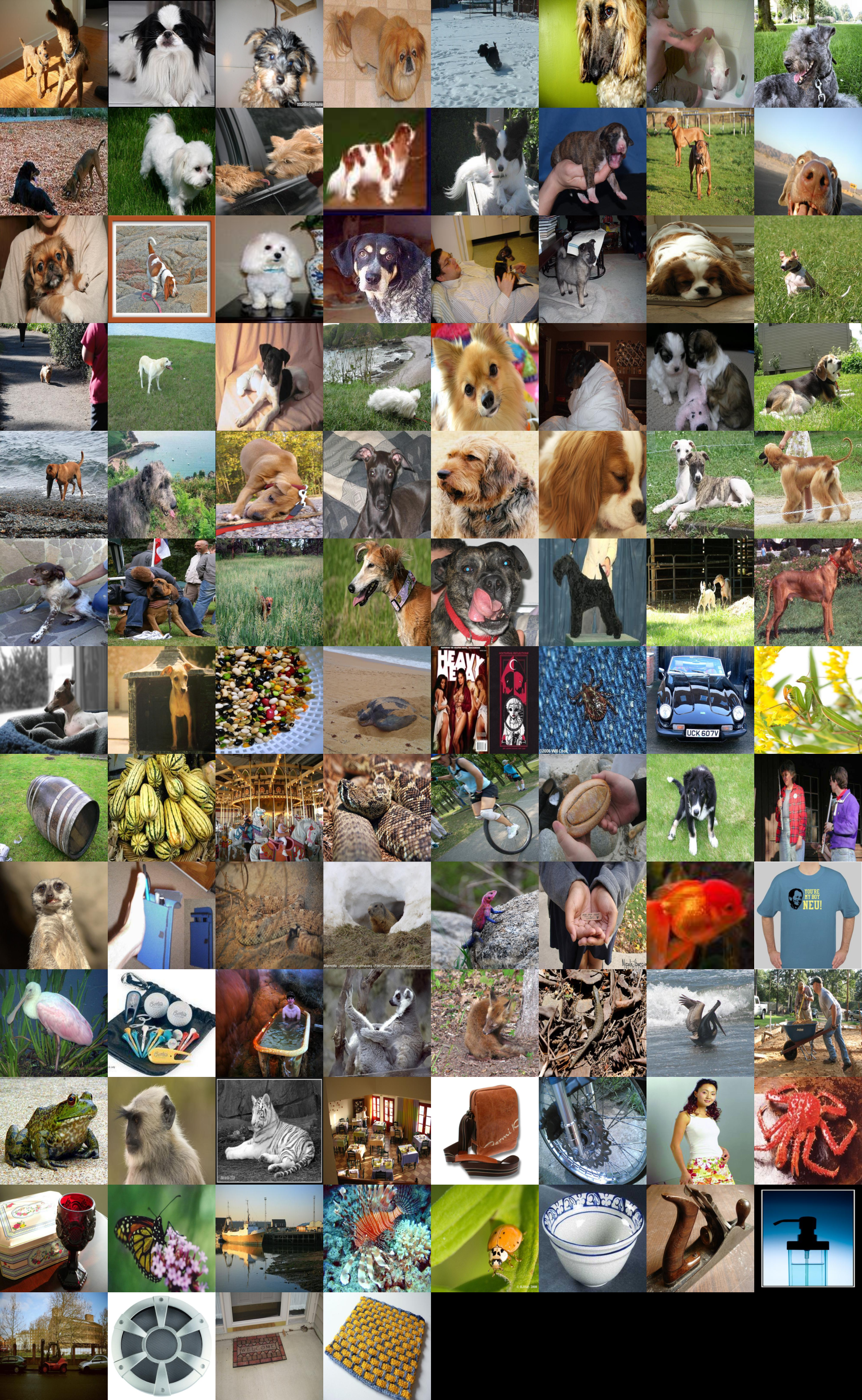}
    \end{minipage}
    \hspace{0.01\linewidth}
    \begin{minipage}[t]{0.45\linewidth}
        \includegraphics[width=\linewidth]{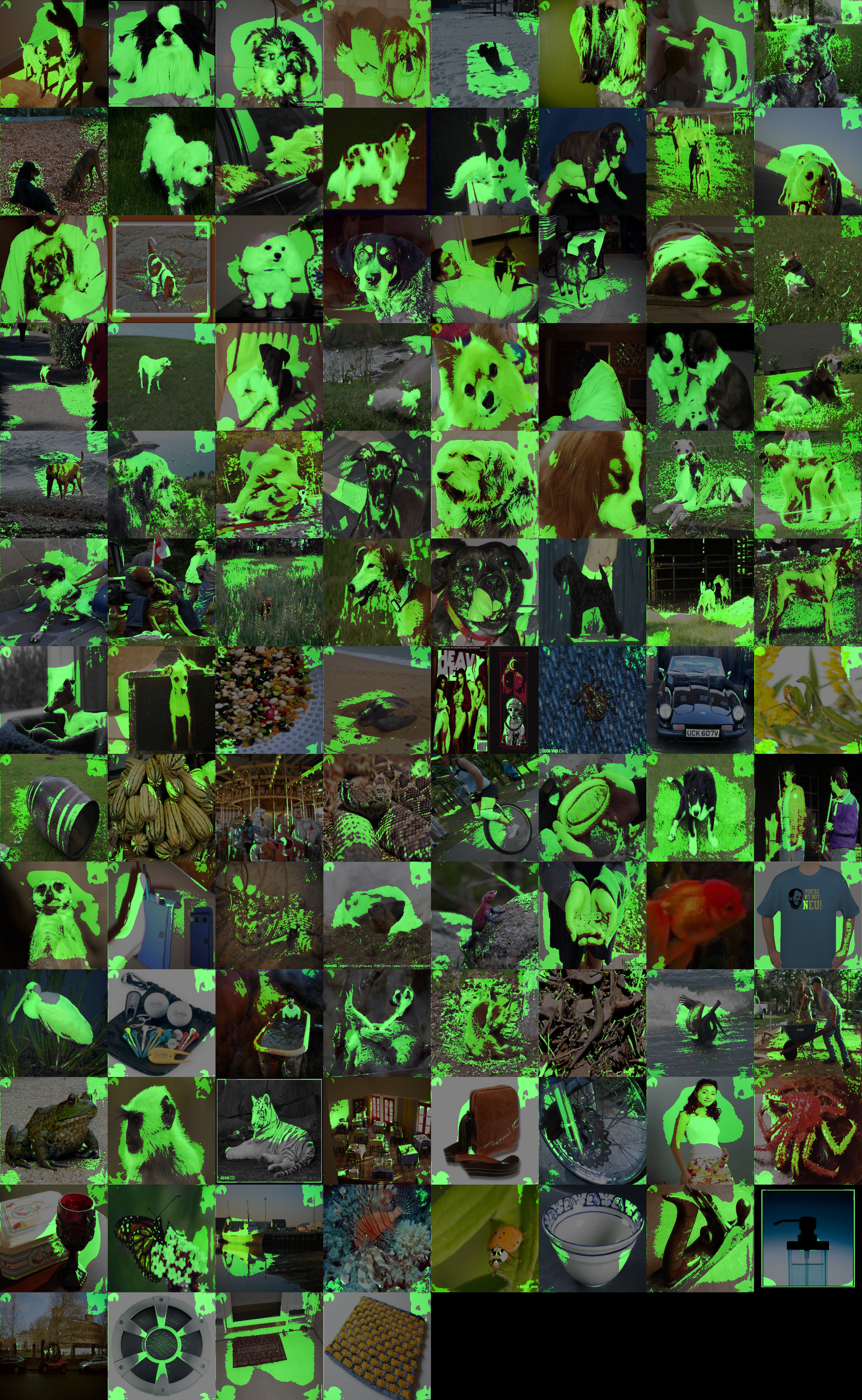}
    \end{minipage}
    \caption{Qualitative results on 50 randomly selected \textbf{validation samples} for "dog-present" images (first 50) and "dog-absent" (last 50) images respectively. Left, each square: input image. Right, each square: output mask overlayed on the input image.}
    \label{fig:val_qua_results}
\end{figure}

Figure \ref{fig:train_qua_results} and Figure \ref{fig:val_qua_results} illustrates the results we obtained for the training and validation set respectively. We see that our model in many cases produces masks with crisp object boundaries that adheres to low-level image appearances, which is achieved without explicit regularization on the masks' consistency with pixel appearances. More importantly, the model appears to be robust to the imbalance and ambiguity related to the "background" class. Without explicit regularization on the background's mask using prior beliefs on the supposed size of the background nor its appearances, the masks in most cases exhibits a focus on the objects of interest and produces a reasonable mask size for the background.

In regards to failure cases, we observe that the model has a tendency towards false positives in the dog-mask in "dog-absent" images with dog-correlated features, for example, images with the presence of humans. We suggest that this caused by correlated concepts with the objects of interest. In this case, since our training images consist of mostly images of dogs accompanied by humans, the network has learned an undesired correlation of the presence of dogs to the presence of humans, and has reflected that in its mask predictions.

\section{Conclusion}

In this work we present a method for weak segmentation from image tags through reconstruction from structured decomposition. By framing segmentation into a case of reconstruction of the image from its respective mask and image decomposition, more rigorous and robust regularization objectives have been realized, and explicit guidance from an image classifier on the task have been made possible. Results on a toy binary segmentation task suggests that our method encourages the network's optimization towards finding better optima, and helps mitigate the problem of under-constraint and background ambiguity.

\subsection{Future work}

We leave the below list as future work:
\begin{enumerate}
    \item \textit{Test on multi-nary segmentation task}: while results on binary segmentation are appealing, we have yet to apply our method onto the more general and standard, multi-categorical segmentation task.
    
    \item \textit{The use of volumetric priors}: prior knowledge on the mask size of presenting objects can be directly utilized in our method if available. It is worth investigating the achievable gains from doing so.
    
    \item \textit{Explore strategies for removing unwanted correlations}: Disentangling the actual objects of interest from their correlated factors remains challenging due to inherent bias arised from data collection. How to achieve this remains an open research question of its own.
\end{enumerate}

\newpage
\appendix

\section{Network Architecture} \label{app:network_arch}

\begin{table}[h!]
\centering
\begin{tabular}{|c|c|c|c|}
\hline
Layer & Type & Output Shape & Parameters \\
\hline
1 & Input Layer & (3, 224, 224) & 0 \\
2 & ConvBlock(c=64, k=3, s=1) & (64, 112, 112) & 38,720 \\
3 & ConvBlock(c=128, k=3, s=1) & (128, 56, 56) & 221,440 \\
4 & ConvBlock(c=256, k=3, s=1) & (256, 28, 28) & 885,248 \\
5 & ConvBlock(c=512, k=3, s=1) & (512, 14, 14) & 3,539,968 \\
6 & Conv(c=1024, k=3, s=1) + LeakyReLU & (1024, 14, 14) & 4,719,616 \\
\hline
  & & & 9,404,992 \\
\hline
\end{tabular}
\caption{
    Layers of the U-Net encoder that is shared between $f_m$ and $f_x$. The Conv(c, k, s) denotes a regular convolution layer with $c$ output channels, $k$ kernel size, and $s$ stride. Each ConvBlock(c, k, s) denotes a composition of the following layers in sequence: [Conv(c, k, s), LeakyReLU, Conv(c, k, s), LeakyReLU, MaxPool(kernel=2, stride=2)].
}
\label{table:enc_layers}
\end{table}

\begin{table}[h!]
\centering
\begin{tabular}{|c|c|c|c|}
\hline
Layer & Type & Output Shape & Parameters \\
\hline
1 & Input from Encoder & (512, 14, 14) & 0 \\
2 & ConvBlock'(c=512, k=3, s=1) & (512, 28, 28) & 9,438,208 \\
3 & ConvBlock'(c=256, k=3, s=1) & (256, 56, 56) & 2,359,808 \\
4 & ConvBlock'(c=128, k=3, s=1) & (128, 112, 112) & 590,080 \\
5 & ConvBlock'(c=64, k=3, s=1) & (64, 224, 224) & 147,584 \\
6 & Conv(c=64, k=3, s=1) + LeakyReLU & (64, 224, 224) & 36,928 \\
7 & Conv(c=$C_{out}$, k=3, s=1) & ($C_{out}$, 224, 224) & 3,462 \\
\hline
  & & & 12,576,070 \\
\hline
\end{tabular}
\caption{
    Layers of the U-Net decoder for $f_m$ and $f_x$, which are not shared across the two. Each ConvBlock'(c, k, s) denotes a composition of the following layers in sequence: [Conv(c, k, s), LeakyReLU, Upsample(ratio=2), Conv(c, k, s), LeakyReLU]. And $C_{out}$ denotes the number of channels in the final output, which differs between $f_m$ ($C_{out} = K$) and $f_x$ ($C_{out} = 3K$). Feature concatenations occur at the input to each ConvBlock', in accordance to the U-Net architecture.
}
\label{table:dec_layers}
\end{table}

\newpage
\bibliography{references}

\end{document}